\documentclass[conference]{IEEEtran}
\IEEEoverridecommandlockouts
\usepackage{cite}
\usepackage{amsmath,amssymb,amsfonts}
\usepackage{algorithmic}
\usepackage{graphicx}
\usepackage{textcomp}
\usepackage{xcolor}
\def\BibTeX{{\rm B\kern-.05em{\sc i\kern-.025em b}\kern-.08em
    T\kern-.1667em\lower.7ex\hbox{E}\kern-.125emX}}
\begin{document}

\title{The Radar-based Activities of Daily Living (R-ADL) Dataset
\thanks{This work is funded through the National Science Foundation Graduate Research Fellowship Program. The authors would also like to thank Cherish for the use of their Home Hub device and Radar Decoding Model.}
}

\author{\IEEEauthorblockN{Dominique Regli}
\IEEEauthorblockA{\textit{Department of Aeronautics and Astronautics} \\
\textit{Massachusetts Institute of Technology}\\
Cambridge, MA, USA \\
dregli@mit.edu}
\and
\IEEEauthorblockN{Brian C. Williams}
\IEEEauthorblockA{\textit{Department of Aeronautics and Astronautics} \\
\textit{Massachusetts Institute of Technology}\\
Cambridge, MA, USA \\
williams@mit.edu}
}

\maketitle

\begin{abstract}
Monitoring Activities of Daily Living (ADLs) supports safety and well-being for people living at home, especially older adults, and radar sensing is gaining increasing attention in the field of activity recognition for its ability to ensure privacy in the home while also surpassing technical limitations of other sensing modalities. However, radar-based activity recognition remains constrained by the limited availability of large-scale datasets. We introduce the Radar-based Activities of Daily Living (R-ADL) dataset to addresses the gaps identified in existing work: 1) it expands the scale of available data by providing over eight hours of annotated recordings in which a person performs eight distinct activities; 2) it overcomes room configuration limitations by recording from five different locations in a two-bedroom, two-bathroom apartment; and 3) it extends beyond primitive actions to focus on high-level ADLs. In addition to the dataset, this paper introduces practical guidance for future ADL dataset construction. We conducted a detailed analysis of how ADLs are prioritized across academic research, health survey tools, and clinical trials to identify clinically significant ADLs. Furthermore, we provide our experimental design, data collection and annotation strategies, post-processing pipeline for forming the R-ADL dataset from raw data, and baseline results on standard machine learning architectures. The R-ADL dataset serves as a valuable resource for advancing ADL recognition systems as we continue building larger and more diverse datasets.
\end{abstract}. 


\section{Introduction}

Monitoring Activities of Daily Living (ADLs) supports safety and well-being for people living at home, especially older adults, by tracking long term health patterns, detecting sudden risks, and enabling prompt response and remediation from caregivers. Despite these benefits, there is currently no widespread access to in-home activity monitoring, largely due to challenges with procuring privacy-preserving activity data to use in computational modeling. While many sensors have been tested for activity recognition, they face various practical challenges in implementation: RGB cameras cannot see in low light, infrared cameras are blocked by occlusions, wearable devices are cumbersome to hold, and more. Most crucially, some devices collect personally identifiable information that is considered invasive to privacy.

Radar sensing is gaining increasing attention in the field of activity recognition for its ability to overcome cultural barriers around ensuring privacy in the home while also surpassing technical limitations of other sensing modalities. However, radar-based activity recognition remains constrained by the limited availability of large-scale datasets \cite{Luo2024, Savvidou2024}. First, radar-based datasets are often restricted to single room configurations \cite{Shen2024, Khan2025Radar}. This lack of diversity in the training data prevents models from generalizing across different environments, resulting in a substantial decline in performance when evaluated in previously unseen settings \cite{El2025}. Second, the majority of these datasets focus on primitive tasks like human pose estimation or posture recognition, but have yet to extend to recognizing complex ADLs \cite{Ding2022, Guendel2021, Roshandel2025, Cao2023, Lee2023}. There is a significant increase in difficulty from recognizing low-level actions like sitting and walking, to recognizing high-level activities such as eating or sleeping. In order to train models to recognize these activities, we need radar-based datasets in which participants perform ADLs across a diverse set of room configurations.

\begin{figure}
    \centering
    \includegraphics[width=\columnwidth]{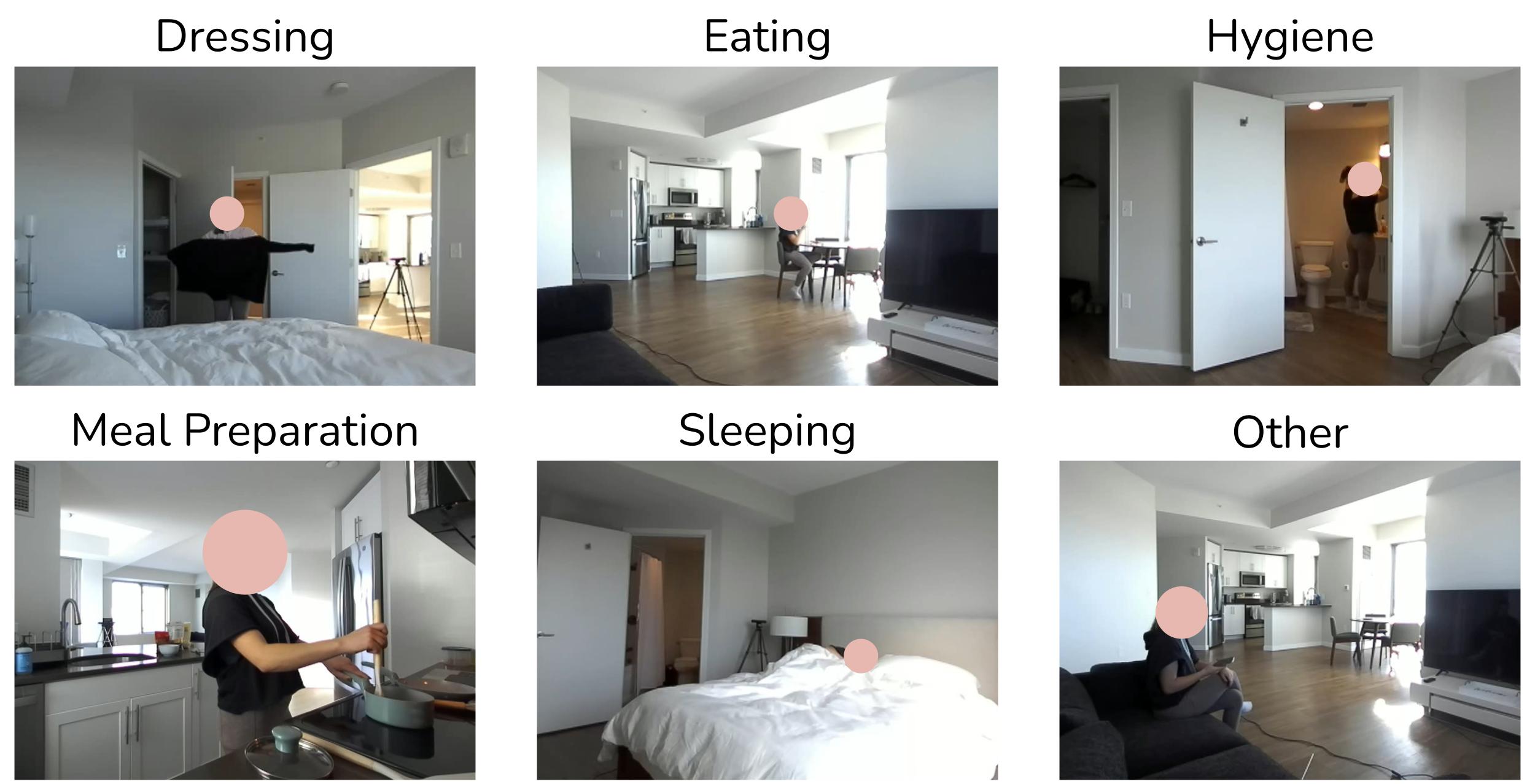}
    \caption{Examples of the ADLs collected in the new R-ADL dataset.}
    \label{ADLExamples}
\end{figure}

We address this gap by introducing the Radar-based Activities of Daily Living (R-ADL) dataset, a novel dataset for radar-based activity recognition research in which a person performs eight distinct activities, examples of which are shown in Figure \ref{ADLExamples}. This dataset addresses the limitations identified in existing work: 1) it expands the scale of available data by providing over eight hours of annotated recordings; 2) it overcomes room configuration limitations by recording from five different locations in a two-bedroom, two-bathroom apartment; and 3) it extends beyond primitive actions to focus on high-level ADLs.

In this paper, we further place emphasis on detailing \textit{how} the data was collected and \textit{why} it was collected in this manner. Our goal is not only to introduce a radar-based dataset for activity recognition, but also to provide practical guidance for future radar ADL data collection and annotation efforts. To this end, we conducted a detailed analysis of how ADLs are prioritized across academic research, health survey tools, and clinical trials to identify the activities most relevant to healthcare and senior living applications. We also documented the complete data collection methodology, including the rationale behind the experimental design and step-by-step collection procedures, to facilitate reproducible dataset development. The approach we describe supports the creation of consistent and clinically relevant radar-based ADL datasets in future research.

The rest of the paper is organized as follows. First, we introduce our analysis of prior surveys to identify clinically significant ADLs and use the results to inform our selection of ADLs in the R-ADL dataset. Next, we detail our experimental design, including factors such as participants, room configurations, and session construction. We then provide our data collection and annotation strategies, and detail how the R-ADL dataset is constructed from raw data. Finally, we show baseline results on standard machine learning architectures to be used as benchmarks for future research.

\section{Identifying Clinically Significant Activities of Daily Living}

This section addresses a central question in developing Activities of Daily Living (ADLs) datasets: what set of activities provides sufficient coverage to support health providers?  There is a non-exhaustive list of activities we do in our home as part of our routine and lifestyle. Ideally, we would like to recognize as many activities as possible, but we need to refine this problem and focus on a subset of activities when collecting data to test novel solution architectures. Therefore, the first critical choice in dataset generation is deciding \textit{which} activities to monitor in order to yield the most relevant information gain.

\subsection{A Multi-Perspective Framework for ADL Analysis}

ADLs are often used in health metrics and survey tools to evaluate a person’s quality of life and ability to provide for themselves, and they can be used over time to assess how someone’s health progressively changes \cite{Pashmdarfard2020}. Since our ultimate goal is to support activity recognition for eldercare and senior living, we scope our analysis towards ADL monitoring in healthcare applications, with particular attention towards healthcare for older adults. With these considerations in mind, we analyze the literature through three complementary lines of evidence:

\begin{enumerate}
    \item \textbf{Academic Research:} In this pillar, we review papers from researchers in academia that introduce ADL datasets for testing new physical devices and computational methods for activity recognition.
    \item \textbf{Health Survey Tools:} This pillar reviews papers that propose new diagnostic metrics that use ADLs when assessing someone's health and well-being. 
    \item \textbf{Clinical Trials:} Here, we identify papers that use ADL survey tools with real patients in live trials to assess trends in their health or monitor their overall well-being. 
\end{enumerate}

Generally, papers introducing new ADL and activity recognition datasets perform literature review based solely on other academic research papers \cite{Diraco2023, Alemdar2013, Intille2005, Tapia2004, Kasteren2008, cumin2017dataset, ordez2013, Arrotta2022, Jiang2026}. However, we argue that it is important to consider at all three perspectives when analyzing which ADLs are most relevant in healthcare. First, as prior work has established, we need to understand what our counterparts in academia have studied to determine what is well known and where there may be gaps in the literature. Second, health survey tools help us understand which ADLs real nurses and doctors think are important. This either reassures us that academics have accurately reflected the clinical priorities, or helps us identify areas that have been understudied in academic work to design ADL monitoring systems.

Finally, reviewing the ADLs used in clinical trials is arguably the most important cornerstone on which to base our research. While tens to hundreds of survey tools exist that integrate ADL assessment, there is a small subset of gold-standard survey tools most commonly applied in clinical trials to assess health for older adults. The ability to monitor these activities --- the ones that are already at the forefront of trials with real people --- will have the most direct impact for healthcare workers and their patients. With these perspectives established, we now introduce our analysis of ADL surveys.

\subsection{Evidence Selection and Data Extraction Strategy}

In order to determine which activities are relevant for healthcare monitoring, we look to prior literature to identify popular activities to monitor and the motivation behind these choices. We note that our goal is not to introduce another survey of ADLs, rather, our contribution lies in analyzing trends across these sources to identify activities that are consistently prioritized for healthcare monitoring. To accomplish this, we build upon existing surveys that have already performed comprehensive literature collection. In particular, we extend work from \cite{Pashmdarfard2020} in surveying ADL assessment tools, and \cite{Diraco2023} in surveying ADL datasets from academia. Collectively, the surveys we draw from consisted of 77 papers in total, with 28 from academia \cite{Diraco2023, Alemdar2013, Intille2005, Tapia2004, Kasteren2008}, 28 from health survey tools \cite{Pashmdarfard2020, Bucks1996, De1998, Spiegel1991, Edemekong2019, Kurlowicz1999, Tierney1988, Berg1989, Brodaty2002, Boop2020, Pan1994, Wenborn2008, Weening2011, Pfeiffer1975, Graessel2009, Dijkstra1999, Davis1997}, and 21 from clinical trials \cite{Den2017, Williams1994, Carpenter2006, Wang2009, Fong2015, Cohen1995, Morris1999, Cremer2023, Ballard2001, Cumming2000, Inagawa2013, Huang2021, Smith1987, Drageset2004, Mlinac2016, Lee2017, Henskens2018, Kruse2013, Marx1992, Laukkanen2001, Burge2013}.

Given this set of papers, we next performed data processing to consolidate naming conventions across papers. There were two primary types of consolidation. First, we grouped synonyms that refer to the same activity. For example, the terms ``Sleep", ``Bed", and ``Nap" are all considered as ``Sleep" in the final dataset. Second, we consolidate instances where activities were referred to with different levels of specificity. Here, activities like ``Eating Soup", ``Consuming Sandwich", ``Lunch", ``Dinner", and ``Eating" are all labeled under the category of ``Eating". These data extraction strategies ensure an accurate representation of the relative frequency of activities across the surveyed datasets.

\subsection{ADL Frequency Analysis and Results}

\begin{figure}
    \centering
    \includegraphics[width=\columnwidth]{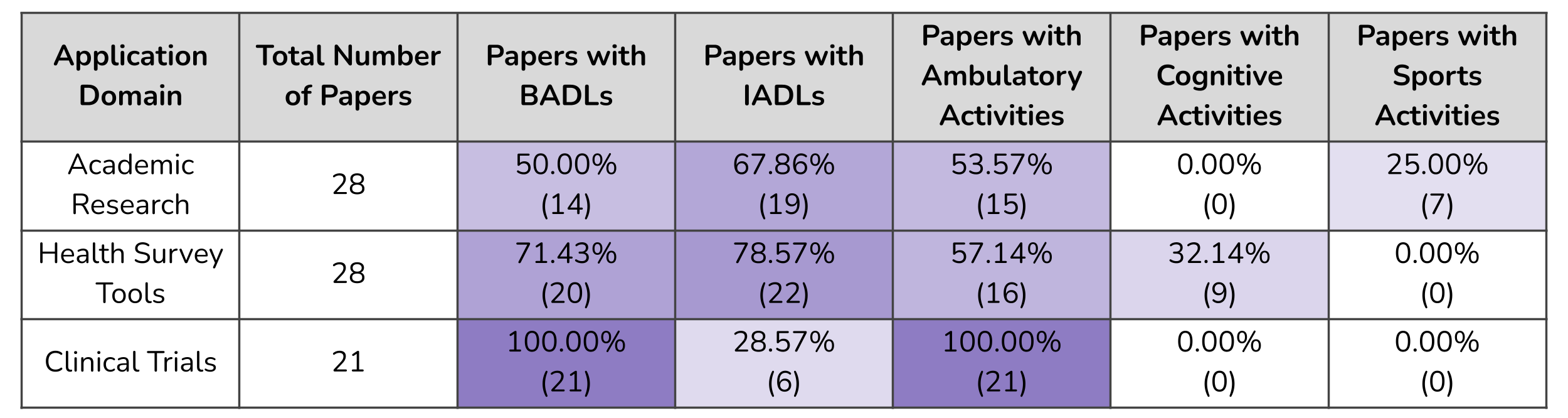}
    \caption{The overall distribution of how frequently each category of activities was included in papers from academia, health survey tools, and clinical trials. Both percentage and total paper counts are included for each category.}
    \label{ADLApplicationDistribution}
\end{figure}

Figure \ref{ADLApplicationDistribution} provides the overall distribution of how frequently each category of activities was included in papers from academia, health survey tools, and clinical trials. At a high level, we categorized the activities in the datasets into five groups: 1) Basic Activities of Daily Living (BADLs), 2) Instrumental Activities of Daily Living (IADLs), 3) ambulatory activities, 4) cognitive activities, and 5) sports activities.

BADLs are the activities required to take care of one's bodily and physical needs, allowing for basic survival and well-being \cite{Edemekong2019}. ADL literature generally agrees that there are six BADL activities: ambulating, continence, dressing, eating, hygiene, and toileting. IADLs, then, are the broader set of skills that can support one's daily life in the home and in community living \cite{Edemekong2019}. These activities can be more cognitively complex and require more planning and socializing. There is a non-exhaustive list of IADLs that are possible in the home, such as meal preparation, laundry, and taking a phone call.

While ambulating is often considered a BADL, we identify it as its own category here to extend the concept to various forms of primitive action and movement. This includes simple walking and transitioning between postures in the home, but it also includes more complex movements like running or carrying objects. Cognitive activities are tasks like memorization and problem solving that are used to assess cognitive function. Finally, multiple ADL datasets specifically called out labels for sports like tennis, soccer, and riding a bicycle.

Taking a closer look at Figure \ref{ADLApplicationDistribution}, there are clear and noticeable trends in the data that help inform which ADLs to monitor in activity recognition research. BADLs are the most commonly integrated into academic research, survey tools, and clinical trials. Collectively, 55 of the papers use BADLs, which is equivalent to 71.43\% of the entire survey. Notably, every single clinical trial used BADLs in their work, emphasizing the importance of BADLs for real-world evaluation of health for older adults. Ambulatory activities follow a similar trend, as they are included in 52 papers or 67.53\% of the overall survey, and once again, every single clinical trial included some form of monitoring for one's ambulatory abilities.

The more surprising trend, then, comes in the IADL category. IADLs are used in 67.86\% of academic papers and 78.57\% of health survey tools, which is more than any of the other four activity categories in both pillars. This could lead us to believe that IADLs are the most important area for research. However, they are only used in 28.57\% of actual clinical trials. Although researchers continue to study IADLs and clinicians have developed numerous IADL-based assessment tools, these activities are incorporated far less frequently the real-world clinical trials. This discrepancy highlights an important gap between clinical trials and the other two perspectives, academic research and survey tools. As a result, IADLs should be interpreted with caution when prioritizing activities for healthcare monitoring, as their prevalence in the literature does not necessarily reflect their use in clinical practice.

Lastly, the cognitive and sports activities were only monitored in single survey pillars: the survey tools and academic research, respectively. While these are both interesting ways of testing one's quality of life, we assert that they are not central to the task of health monitoring in the home.

\begin{table}
\centering
\caption{Activities of Daily Living (ADLs) in the R-ADL Dataset.}
\label{tab:ADLs}

\begin{tabular}{|p{1.3cm}|p{6cm}|}
\hline
\multicolumn{1}{|c|}{\textbf{ADL}} &
\multicolumn{1}{c|}{\textbf{Definition}} \\ \hline

Ambulating & The ability to sit, stand, lie down, or ambulate while not performing a specific ADL \\ \hline

Bathing & The ability to clean one's body with soap and water \\ \hline

Dressing & The ability to get clothes and put them on one's body \\ \hline

Eating & The ability to eat food and feed oneself by getting food from a dish to one's mouth \\ \hline

Hygiene & The ability to perform grooming tasks such as dental care, skin care, or hair brushing \\ \hline

Meal Preparation & The ability to organize and cook food so that it is ready to eat \\ \hline

Toileting & The ability to get to and from the toilet, use it appropriately, and clean oneself after \\ \hline

Sleeping & The ability to lie down and fall asleep \\ \hline

\end{tabular}
\end{table}

\subsection{Selection of ADLs for the R-ADL Dataset}

Motivated by our analysis of ADL frequency across the multi-perspective framework, we selected the ADL outputs defined in Table \ref{tab:ADLs} for data collection in the R-ADL dataset. We include the fundamental BADLs, as well as ambulation, sleeping, and meal preparation. It is clear from Figure \ref{ADLApplicationDistribution} that BADLs and ambulatory movement are most the pivotal to real-world clinical trials, and are still of high interest in academic research and health survey tools. Our dataset captures all BADLs excluding continence, which was determined to be impractical for activity recognition. We include the IADL of Meal Preparation because it was the most frequent ADL to be included in survey tools, on par with BADL frequency. Finally, we include the activity of sleeping. Sleeping has recently been reclassified from an ADL to an occupational need by the American Occupational Therapy Association because of how essential it is for regulating physiological functioning \cite{Tester2018}. While not strictly an ADL, we argue that the reclassification only emphasizes the importance of recognizing sleep in activity monitoring, and therefore we include it in the dataset. Together, this set of ADLs provides coverage in supporting healthcare providers through activity recognition.

\section{Experimental Design}

The goal in our experimental design for the R-ADL dataset collection was to optimize for the size of dataset, the quality of data collected, and the diversity of the scenes being recorded. Practical and logistical considerations were also weighed in decision making when designing the collection setup. Together, these objectives and real-world conditions informed a number of features of the dataset, which we now explore.

\subsection{Activities}
Once we selected the activities in Table \ref{tab:ADLs}, we had to determine how the activities would be described to participants being recorded in data collection sessions. In our setup, a volunteer acts out the performance of the activities as they would in day-to-day life. Therefore, we need to give them a general idea of what each activity involves doing, without telling them explicitly how to do it. 

In line with prior work, we wrote high-level scripts for each ADL to balance between establishing common definitions of activity types and variety in how these activities are performed \cite{cumin2017dataset, ordez2013, Arrotta2022, Jiang2026}. These scripts include where the person is expected to be, how they are expected to move, what actions they can perform, and what props they can use in the process. We include example scripts for sleeping and dressing activities. These scripts demonstrate how there are typical features of each activity that we expect to occur universally, but there are also less common aspects of the activities that we still wish to capture in the data. Some activities are more standard, like sleeping, and have relatively basic instructions. Meanwhile, an activity like dressing can be performed in many different ways, and so the script encourages more variety.

\vspace{2pt}

\noindent \underline{Sleeping Script}
\begin{itemize}
    \setlength{\itemsep}{2pt}
    \setlength{\parskip}{0pt}
    \setlength{\parsep}{0pt}

    \item Location: Participant is physically in a bed.
    \item State: Participant is lying down.
    \item Props: Pillows, Covers
    \item Prompts:
    \begin{itemize}
        \setlength{\itemsep}{1pt}
        \setlength{\parskip}{0pt}
        \setlength{\parsep}{0pt}
        \item Must emulate lying down asleep
        \item May rotate or roll over, but not often
    \end{itemize}
\end{itemize}

\vspace{2pt}

\noindent \underline{Dressing Script}
\begin{itemize}
    \setlength{\itemsep}{2pt}
    \setlength{\parskip}{0pt}
    \setlength{\parsep}{0pt}

    \item Location: Participant is usually in the walk-in closet and bedroom. They can move between both spaces. They can also go into the bathroom to check appearance.
    \item State: Participant is in any state besides lying down, and there tends to be lots of movement as they change and check how an outfit looks.
    \item Props: Three Shirts, Three Pairs of Pants, One Cardigan, Two Pairs of Socks, Two Pairs of Shoes
    
    \item Prompts:
    \begin{itemize}
        \setlength{\itemsep}{1pt}
        \setlength{\parskip}{0pt}
        \setlength{\parsep}{0pt}
        \item Must emulate putting on pants, a shirt, and shoes
        \item May use any of the props available to them
        \item May do these actions in any order and repeat actions
        \item May emulate putting on other pieces of clothing outside of those required, like socks or a cardigan
        \item May emulate switching between outfits
    \end{itemize}
\end{itemize}

\subsection{Participants}

The R-ADL dataset recorded a single participant for the full eight hours of data collection. Future work should expand this dataset to include additional participants, with particular focus on recruiting older adults, but we assert that the lack of variety in participants is not a crucial limitation for the purpose of evaluating approaches to activity recognition. Results from prior work indicate that diversity in room configurations is the dominant source of model overfitting for radar-based ADL monitoring, whereas variation in participants' body proportions has a comparatively minor effect \cite{El2025}.

\begin{figure}
    \centering
    \includegraphics[width=\columnwidth]{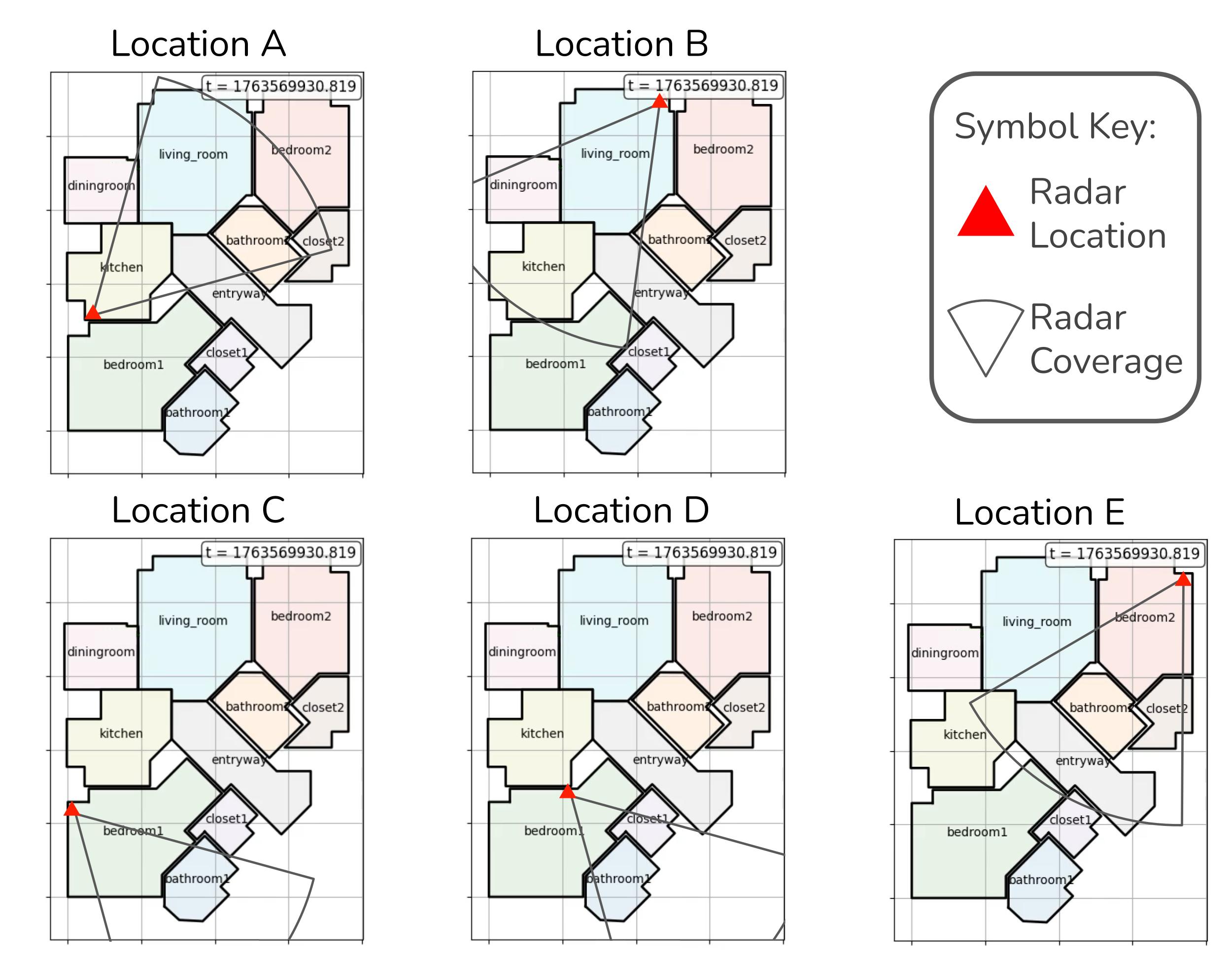}
    \caption{The five apartment locations used for data collection.}
    \label{ApartmentDataConfigurations}
\end{figure}

\subsection{Floor Plan}
We used a two-bedroom, two-bathroom apartment for data collection. The common areas include a kitchen, dining room, living room, and entryway; there are also two walk-in closets. In total, we collected data at five different locations in the apartment, shown in Figure \ref{ApartmentDataConfigurations}. Providing a variety of locations is pivotal to evaluating how well models can generalize to perform ADL classification in unseen environments. The selection of these five locations was focused on optimizing variation within the space provided, while avoiding unnecessary overlap in data collection configurations. For example, locations A and B monitor the same rooms, but they do so from vastly different perspective angles. This was intentionally done to create diversity in the data available for training and testing.

\subsection{Sessions}

We broke the eight hours of data collection into 30 minute sessions, for 16 sessions in total. In each session, the participant did three different ADLs for 10 minutes each. The sessions were designed such that a participant would be engaged in doing the activity and have ample time to perform it in a variety of ways, without sessions being so long that the participant would become tired or distracted in the process. Breaks were provided between sessions of data collection to support the attention span and energy levels of the participant, further ensuring high quality of the recorded data.

The full list of sessions is provided in Table \ref{tab:ADLSessions}. Each activity was performed six times across different sessions, resulting in a total of one hour of data collection per activity. Three data sessions were recorded from each of the locations A-D, for a total of 1.5 hours of data in each of the radar configurations. Location E was used a fourth time to get an additional 30 minutes of data for the second bedroom. Activities were assigned to sessions in order to maximize heterogeneity, mixing the order of activities performed, the combination of activities in each session, and the location where the activities are performed. This entropy in the dataset is only constrained by which activities can be performed in which room locations. For example, a bathing activity would not be executed in common areas such as the kitchen or the living room, so it is not recorded from locations A and B.

\begin{table}
\centering
\caption{The data collection sessions for the R-ADL dataset .}
\label{tab:ADLSessions}

\begin{tabular}{|c|c|c|}
\hline
\multicolumn{1}{|c|}{\textbf{Session Number}} &
\multicolumn{1}{c|}{\textbf{Location}} & 
\multicolumn{1}{c|}{\textbf{Activities}} \\ \hline

1 & A & Meal Preparation, Eating, Ambulating \\ \hline

2 & A & Eating, Ambulating, Meal Preparation \\ \hline

3 & A & Meal Preparation, Ambulating, Eating \\ \hline

4 & B & Ambulating, Eating, Meal Preparation \\ \hline

5 & B & Meal Preparation, Eating, Ambulating \\ \hline

6 & B & Eating, Meal Preparation,Ambulating \\ \hline

7 & C & Sleeping, Toileting, Hygiene \\ \hline

8 & C & Bathing, Toileting, Sleeping \\ \hline

9 & C & Hygiene, Dressing, Sleeping \\ \hline

10 & D & Bathing, Hygiene, Dressing \\ \hline

11 & D & Toileting, Bathing, Hygiene \\ \hline

12 & D & Dressing, Bathing, Toileting \\ \hline

13 & E & Dressing, Toileting, Sleeping \\ \hline

14 & E & Hygiene, Bathing, Dressing \\ \hline

15 & E & Sleeping, Toileting, Hygiene \\ \hline

16 & E & Bathing, Dressing, Sleeping \\ \hline

\end{tabular}
\end{table}

\section{Data Collection and Annotation}

\subsection{Data Collection}

\begin{figure}
    \centering
    \includegraphics[width=0.7\columnwidth]{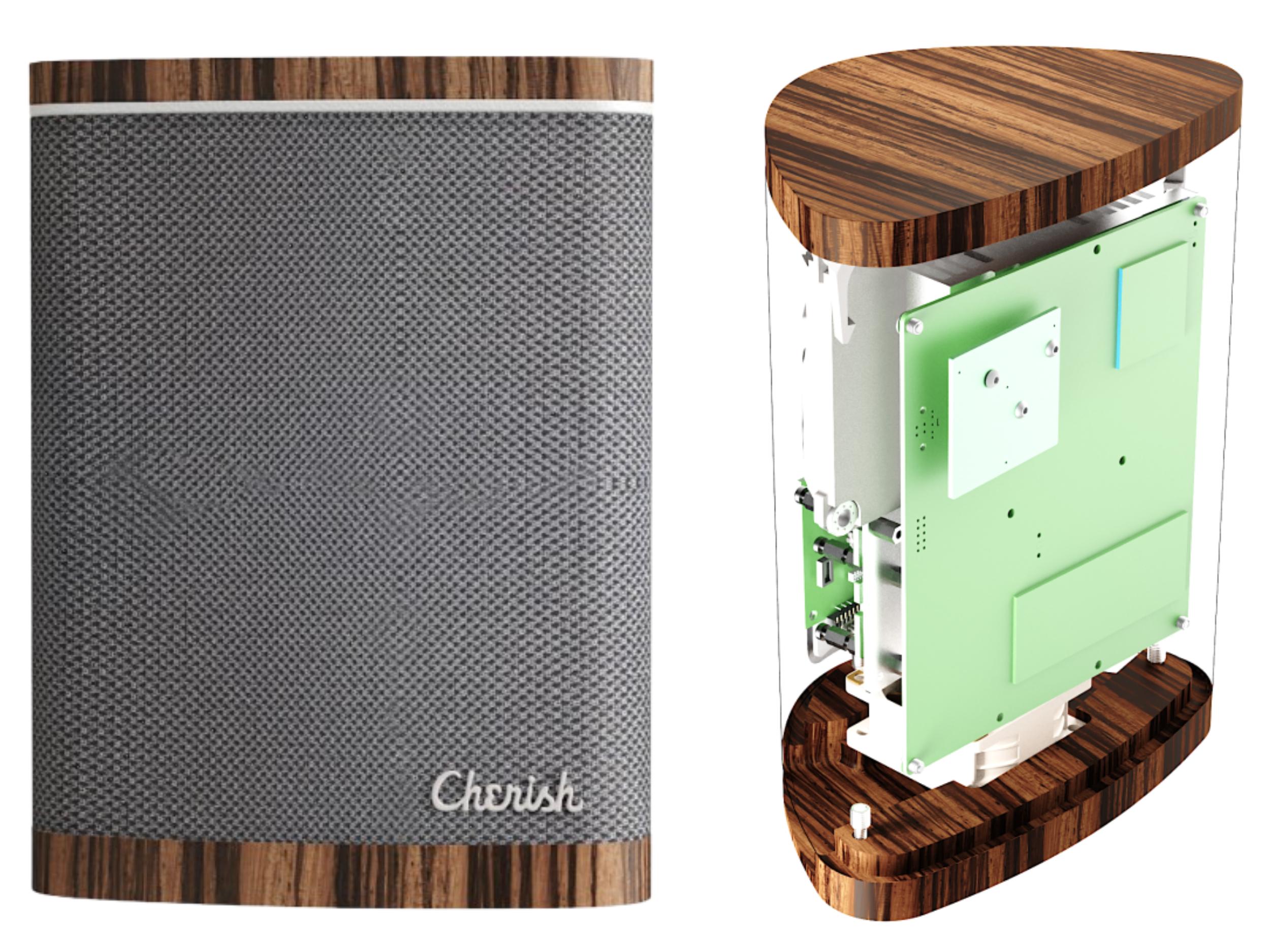}
    \caption{The Cherish Home Hub device (left) and the internals of the device, including the radar board (right).}
    \label{CherishHomeHub}
\end{figure}

\begin{figure}
    \centering
    \includegraphics[width=\columnwidth]{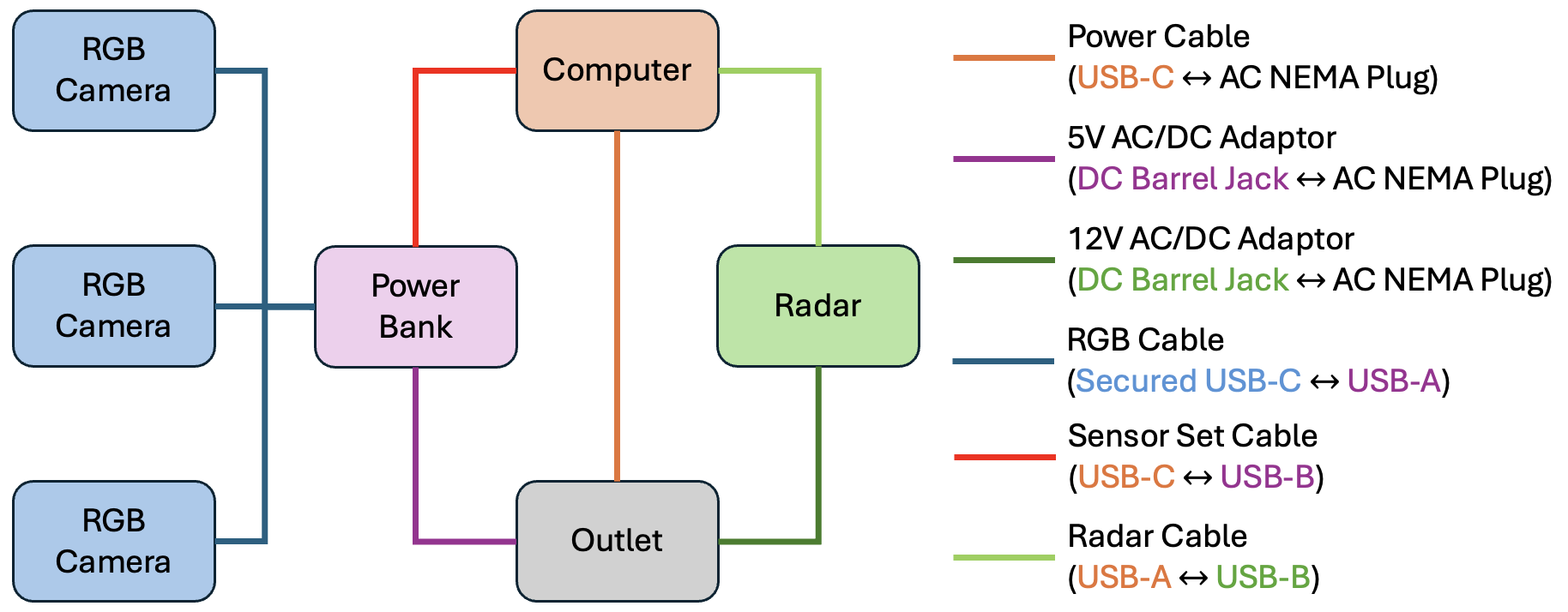}
    \caption{The wiring diagram for experimental setup.}
    \label{WiringDiagram}
\end{figure}

The full data collection process includes setting up equipment, recording data, and returning everything to its original state at the end of collection. For the hardware configuration, we use a radar transmitter and receiver designed by Cherish. Their device is called the Home Hub, and both the external structure and internals of the radar board are shown in Figure \ref{CherishHomeHub}. In addition to the radar, three RGB cameras recorded video of the rooms to enable data annotation. Figure \ref{WiringDiagram} shows the wiring diagram for equipment setup. The radar and cameras are connected to the same laptop, with each device recording timestamped data to synchronize the sensor streams.

Next, data was recorded using the following checklist, which outlines the exact steps taken to support quality assurance, continuity across sessions, and efficient data collection.

\begin{enumerate}
    \item Follow the wiring diagram to connect the laptop, Home Hub, and three RGB cameras so that all have power and are in communication with each other.
    \item Once connected, position the radar in the designated location (A-E) for the current session. Take manual measurements of the radar location based on height from the floor and distance from a known position along the wall, such as a corner, to confirm the correct positioning. Take pictures of these measurements for continuity.
    \item Place the cameras in their designated locations. This can be approximate, as they are only used for visual labeling.
    \item Run an initial 30 second data collection and inspect the saved data files to confirm everything is set up correctly.
    \item Place the props for the session so that the participant can naturally pick them up during data collection.
    \item Begin the data collection session. The participant performs the activities as specified by their scripts.
    \item Once the 30-minute session is finished, go to the laptop and end the session. Save the data and upload it at the end of the day once all sessions have been completed.
\end{enumerate}

\subsection{Data Annotation}

\begin{figure}
    \centering
    \includegraphics[width=0.5\columnwidth]{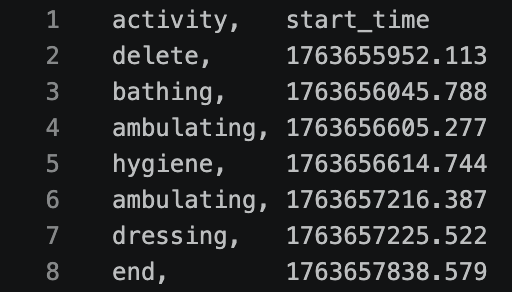}
    \caption{An example of a data annotation file, showing Session 10 annotations. Start time is reported as a Unix timestamp, in seconds, derived from the synchronized RGB camera recording.}
    \label{DataAnnotation}
\end{figure}

Once the data was collected, ground truth activity labels were assigned to the frames of radar data so that computational methods can be evaluated by comparing predictive outputs to true ADL classes. These labels were created manually by observing the video data and recording the timestamps of activities. Reviewers noted each activity's label and start  time, and then the end time was computed internally as the timestamp prior to the start time of the following activity. If anything went wrong during data collection or if there was a segment we did not want to include in the final dataset, this was labeled as ``Delete". A common case for using ``Delete" was to remove the very beginning of a session when the radar unit was being set up and the participant was getting into position. Finally, the label ``End" was used to denote the termination of the session. An example data annotation file, showing annotations from Session 10, is provided in Figure \ref{DataAnnotation}.

\section{R-ADL Dataset Construction}

To construct the final R-ADL dataset, we perform data processing operations on the raw radar data to create meaningful features for ADL prediction. Radar data is inherently noisy and sparse, so in order to extract accurate and relevant insights from the raw data, we introduce a set of features that describe individuals based on their body posture, movement, and position. Furthermore, we employ qualitative methods to contextualize the person within their home, relating them to key objects in the space. A summary of the data types included in the R-ADL dataset is shown in Table \ref{datatypes}.

\begin{table}
\centering
\caption{Types of data in the R-ADL Dataset.}
\label{datatypes}

\begin{tabular}{|p{1.25cm}|p{1.25cm}|p{1.25cm}|p{2cm}|p{2cm}|}
\hline
\multicolumn{1}{|c|}{\textbf{Name}} & \multicolumn{1}{|c|}{\textbf{Variable}} & \multicolumn{1}{|c|}{\textbf{Format}} & \multicolumn{1}{|c|}{\textbf{Units}} & \multicolumn{1}{|c|}{\textbf{Example}}\\ \hline

Session ID & \multicolumn{1}{|c|}{\textit{session}} &  String & \multicolumn{1}{|c|}{unitless} &
\multicolumn{1}{|c|}{session11}
\\ \hline

Target ID & \multicolumn{1}{|c|}{\textit{target}} &  String & \multicolumn{1}{|c|}{unitless} &
\multicolumn{1}{|c|}{person1}
\\ \hline

Frame & \multicolumn{1}{|c|}{\textit{frame}} &  Integer & \multicolumn{1}{|c|}{unitless} &
\multicolumn{1}{|c|}{247}
\\ \hline

Time & \multicolumn{1}{|c|}{\textit{t}} &  Float & \multicolumn{1}{|c|}{seconds} &
\multicolumn{1}{|c|}{63581144.65}
\\ \hline

Radar\newline Location & \multicolumn{1}{|c|}{\textit{rad}} & Discrete Label & \multicolumn{1}{|c|}{unitless} &
\multicolumn{1}{|c|}{A}
\\ \hline

Centroid & \multicolumn{1}{|c|}{$(x,y)$} & Coordinate Pair & \multicolumn{1}{|c|}{$(m, m)$} &
\multicolumn{1}{|c|}{$(5.78, 9.23)$}
\\ \hline

Body\newline Posture & \multicolumn{1}{|c|}{\textit{bp}} & Discrete Label & \multicolumn{1}{|c|}{unitless} &
\multicolumn{1}{|c|}{Sitting}
\\ \hline

Movement & \multicolumn{1}{|c|}{\textit{mv}} & Discrete Label & \multicolumn{1}{|c|}{unitless} &
\multicolumn{1}{|c|}{Stationary}
\\ \hline

Room\newline Location & \multicolumn{1}{|c|}{\textit{rl}} & Discrete Label & \multicolumn{1}{|c|}{unitless} &
\multicolumn{1}{|c|}{Kitchen}
\\ \hline

Key Object Relations & \multicolumn{1}{|c|}{\textit{obj}} & Discrete Label per Object & \multicolumn{1}{|c|}{unitless} &
Bed: [Disconnected, Far, Stationary] Refrigerator: [Overlap, Close, Towards]
\\ \hline

Activity & \multicolumn{1}{|c|}{$a_{pred}$} & Discrete Label & \multicolumn{1}{|c|}{unitless} &
\multicolumn{1}{|c|}{Eating}
\\ \hline

\end{tabular}
\end{table}

\begin{figure}
    \centering
    \includegraphics[width=\columnwidth]{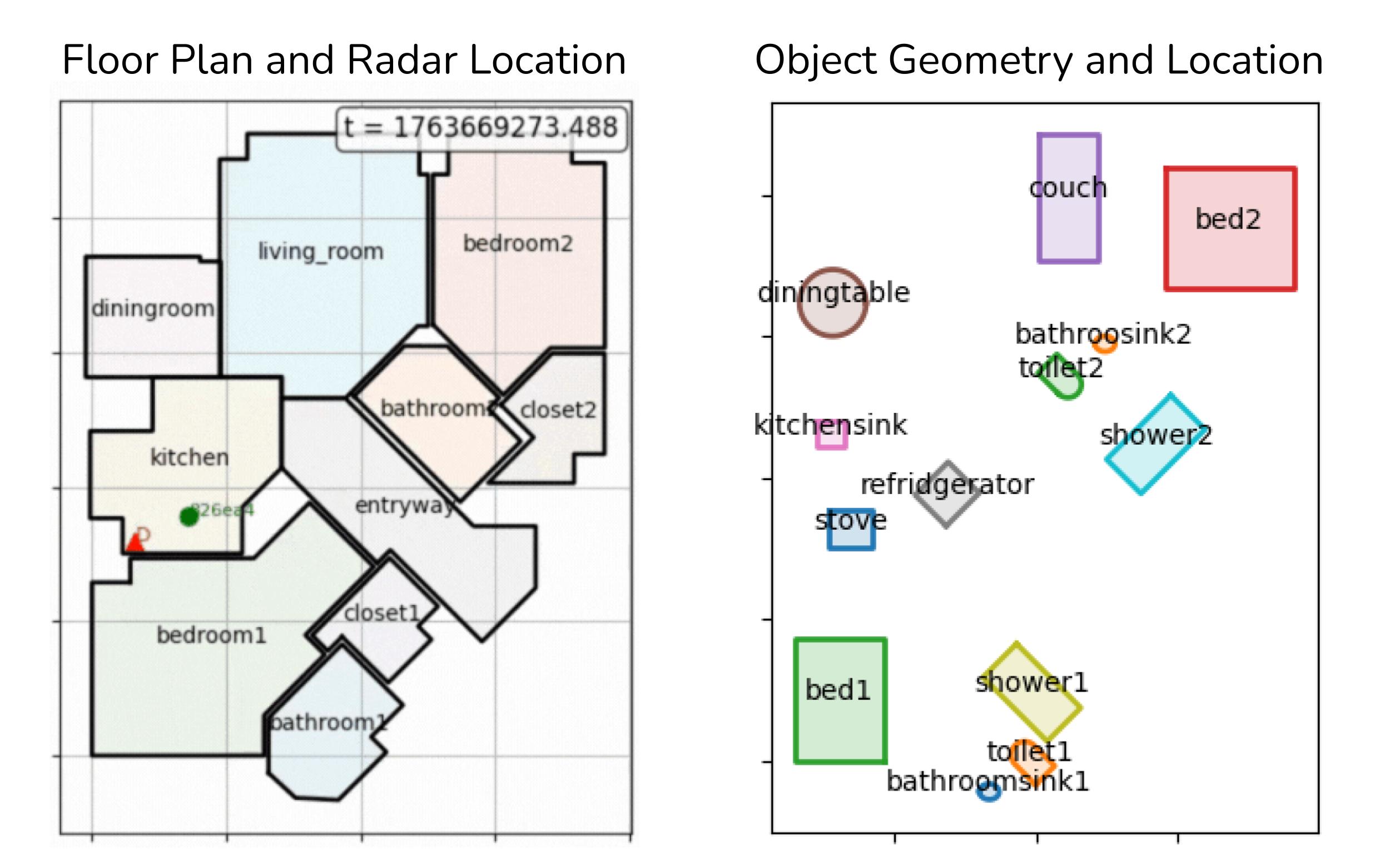}
    \caption{The scene context for the apartment used in data collection. The left image includes the floor plan of the apartment, the location of the radar unit as a red triangle, and a localized target as a green circle. The right image shows the 2D shape of each object's geometry relative to the floor plan.}
    \label{SceneContext}
\end{figure}

Crucial to the construction of the R-ADL dataset is access to spatial information about the floor plan and key objects in the apartment. In addition to the recorded data, R-ADL provides a floor plan file, which includes 2D spatial data for the size and shape of the rooms being monitored, as well as labels for each room type. Similarly, the object file includes each object's physical shape, label, and position within the floor plan. Figure \ref{SceneContext} summarizes this scene context for the apartment used in data collection. With this scene context established, the rest of this section reviews each feature included in the R-ADL dataset and how it was constructed. 

\subsection{Metadata}

The metadata at each data point tracks general purpose information to support dataset organization. For example, we include which session the data point was collected in and the radar location in that session. Within sessions, we also track exact timestamps and consecutive frame counts. Finally, we save the target ID, which will be important for differentiating between people when there is more than one person detected.

\subsection{Centroid}

Centroid data consists of a person's 2D position in a bird's-eye view of the floor plan. These coordinates are provided both in a local frame of reference relative to the radar unit, and a global frame of reference relative to the floor plan diagram. The centroids are extracted using the Radar Decoding Model from Cherish, which leverages large-scale machine learning to convert raw radar signals into localized centroid detections. This model follows established approaches from prior work and is not a novel contribution of this paper, but it provides a standardized spatial representation that enables the dataset to support downstream ADL recognition and analysis.

\subsection{Body Posture}

Body posture data provides a prediction of whether a person is standing, sitting, or lying down. It also includes the underlying probability distribution over the three posture classes, from which the maximum likelihood candidate is selected. This enables prediction models to either use discrete posture labels or consider probabilistic weighting in activity recognition. As with centroid data, this information is extracted using the Radar Decoding Model.

\subsection{Movement}

Movement identifies whether a person is moving or stationary at a given timestamp. Since people will not be perfectly still in real world data, a person is labeled as stationary if their centroid position falls within an error margin of minimal movement across sequential time steps. This calculation is shown in Equation \ref{mv_calc} using euclidean distance for error margin $\epsilon= 0.51 cm \text{ } (0.2 in)$ at timestep $t$.
\begin{equation}
\label{mv_calc}
mv_t =
\begin{cases}
\text{stationary}, & \|(x_{t}, y_{t})-(x_{t-1}, y_{t-1})\|_2 \leq \epsilon \\
\text{moving}, & \text{otherwise}
\end{cases}
\end{equation}

\subsection{Room Location}

Room location provides the label for the room the person is in, based on their centroid position and the floor plan. Each room, $r$, is treated as a polygon, $\mathcal{P}_r$, and \texttt{shapely} is used to determine the room location as shown in Equation \ref{rm_calc}.
\begin{equation}
\label{rm_calc}
rl_t = r \in Rooms \;\; \text{where} \;\; (x_t, y_t) \in \mathcal{P}_r
\end{equation}

\subsection{Key Object Relations}

\begin{figure}
    \centering
    \includegraphics[width=\columnwidth]{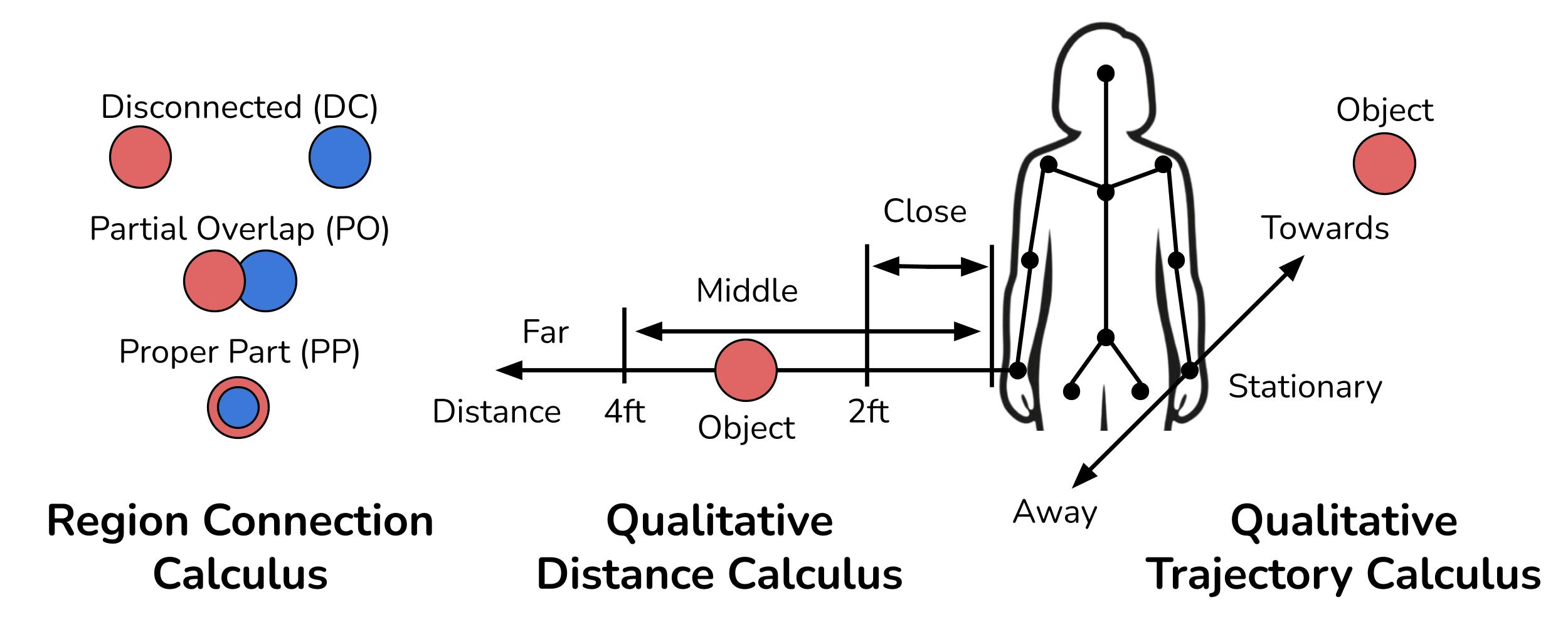}
    \caption{A visual representation of RCC, QDC, and QTC.}
    \label{QualitativeSpatialCalculi}
\end{figure}

To further support ADL prediction, we leverage the field of Qualitative Spatial Reasoning (QSR) to relate the person to key objects within their home. QSR methods focus on mathematical principles of spatial relations between objects and how these relations can be described in a qualitative manner \cite{CohnQSR}. This approach builds off human-intuitive and semantic knowledge of the physical world in order to create common-sense abstractions that are relevant to the computational tasks we wish to model. For each key object, we use three different relations to qualitatively describe the object in relation to the human: Region Connection Calculus (RCC), Qualitative Distance Calculus (QDC), and Qualitative Trajectory Calculus (QTC). These relations describe connectedness, distance, and motion, respectively. The QSR relations are summarized in Figure \ref{QualitativeSpatialCalculi}, and we next describe how each is computed.

First, RCC describes how two objects relate in terms of spatial connectivity, including boundary, intersection, and enclosure between objects \cite{Cohn1997}. We define three RCC state assignments, where the human is either disconnected from, partially overlapping with, or a proper part of a key object. We use \texttt{shapely} to calculate our RCC values as shown in Equation \ref{rcc_calc}, where $\mathcal{P}_{H}$ and $\mathcal{P}_{O}$ are the polygons for the human, $H$ and each key object, $O$.
\begin{equation}
\label{rcc_calc}
\text{RCC}(H_t,O)_t =\\
\begin{cases}
\text{Disconnected}, & \mathcal{P}_{H,t}\cap\mathcal{P}_{O}=\emptyset\\
\text{Proper Part}, & \mathcal{P}_{H,t}\subset\mathcal{P}_{O}\\
\text{Partial Overlap}, & \text{otherwise}
\end{cases}
\end{equation}

Next, QDC discretizes the physical distance between objects into qualitative categories \cite{Hernandez1995}. We include three distance ranges for the R-ADL dataset. Equation \ref{qdc_calc} shows our instantiation of QDC with our manually defined thresholds of $r_{qdc\text{–}1} = 0.61m \text{ } (2ft)$ and $r_{qdc\text{–}2} = 1.22m \text{ } (4ft)$. We use $d_t$ to represent the euclidean distance between the centroids of the object and the person at timestep $t$.
\begin{equation}
\label{qdc_calc}
\text{QDC}(H_t, O)_t =
\begin{cases}
\text{Close}, & d_t \leq r_{qdc\text{–}1}\\
\text{Middle}, & r_{qdc\text{–}1} < d_t \leq r_{qdc\text{–}2} \\
\text{Far}, & d_t >r_{qdc\text{–}2}
\end{cases}
\end{equation}

Finally, QTC represents the relative movement of one object with respect to another \cite{Van2006}. As with the movement feature, we include an error margin of $\epsilon=0.51 cm \text{ } (0.2 in)$ for the stationary label in QTC. The computation of feature assignments is summarized by Equation \ref{qtc_calc}.
\begin{equation}
\label{qtc_calc}
\text{QTC}(H_t, O)_t =
\begin{cases}
\text{Stationary}, & |d_t - d_{t-1}| \leq \epsilon \\
\text{Towards}, & d_t - d_{t-1} < -\epsilon \\
\text{Away}, & d_t - d_{t-1} > \epsilon
\end{cases}
\end{equation}

\begin{figure}
    \centering
    \includegraphics[width=\columnwidth]{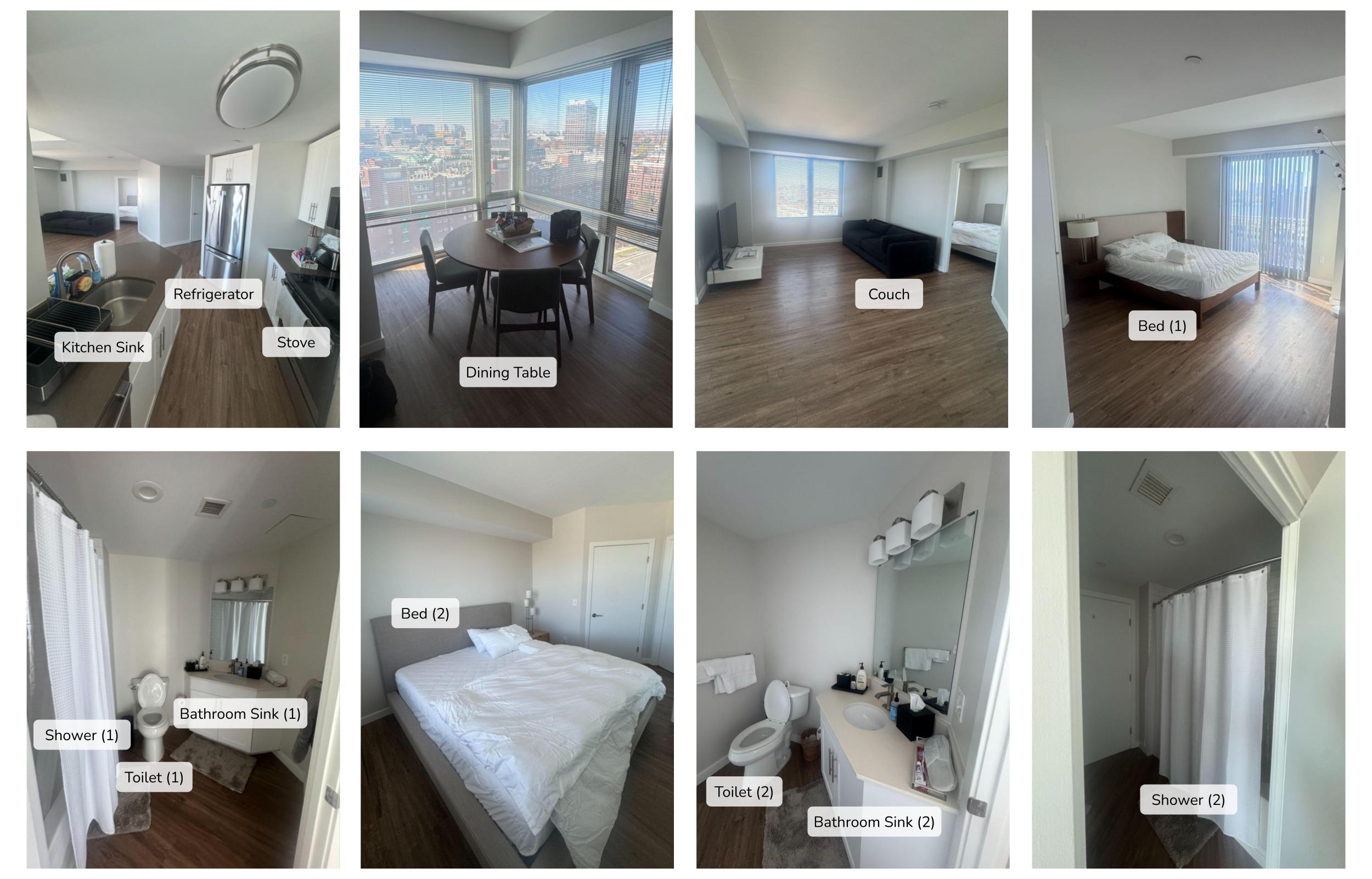}
    \caption{The key objects in the apartment used for ADL data collection.}
    \label{KeyObjects}
\end{figure}

When implementing QSR relations for the R-ADL dataset, we selected nine key household objects: bathroom sink, bed, couch, dining table, kitchen sink, refrigerator, shower, stove, and toilet. Figure \ref{KeyObjects} shows these objects for the apartment used in data collection. While there are many household objects we could have used in these qualitative descriptions, we identified three driving requirements for selecting key objects in the home. First, these objects need to be standard items commonly found in every home to ensure our dataset is generalizable across diverse households. Second, the objects we use must be stationary in the home. The scene context is input offline to the model and is independent from the real-time radar data; therefore, we need objects that have fixed locations regardless of how the human interacts with their environment. Finally, the objects must be relevant to the task of ADL classification. Every house tends to have a thermostat in a fixed location, but the thermostat is irrelevant when trying to determine whether a person is eating or sleeping. The objects must be ones that people interact with when performing activities, such that they are probabilistically correlated with certain activities and can be used in ADL inference. 

\section{Baseline Evaluation of ADL Classification on the R-ADL Dataset}

In this section, we establish baseline performance for ADL classification on the R-ADL dataset using representative machine learning models. These baselines provide a standardized point of comparison for future work, enabling researchers to test new methods under a common evaluation protocol and quantify improvements in performance on the dataset.

\subsection{Model Architectures}

We introduce two models with distinct architectures and unique strengths for baseline evaluation, namely the random forest and neural network methods. Random forests learn by combining many simple decision trees, each trained on different subsets of data, allowing the model to capture structured patterns in classification \cite{Breiman2001}. Neural networks, meanwhile, learn by optimizing layered representations of the data in order to model complex, nonlinear relationships \cite{Schmidhuber2015}. 

\begin{figure}
    \centering
    \includegraphics[width=0.6\columnwidth]{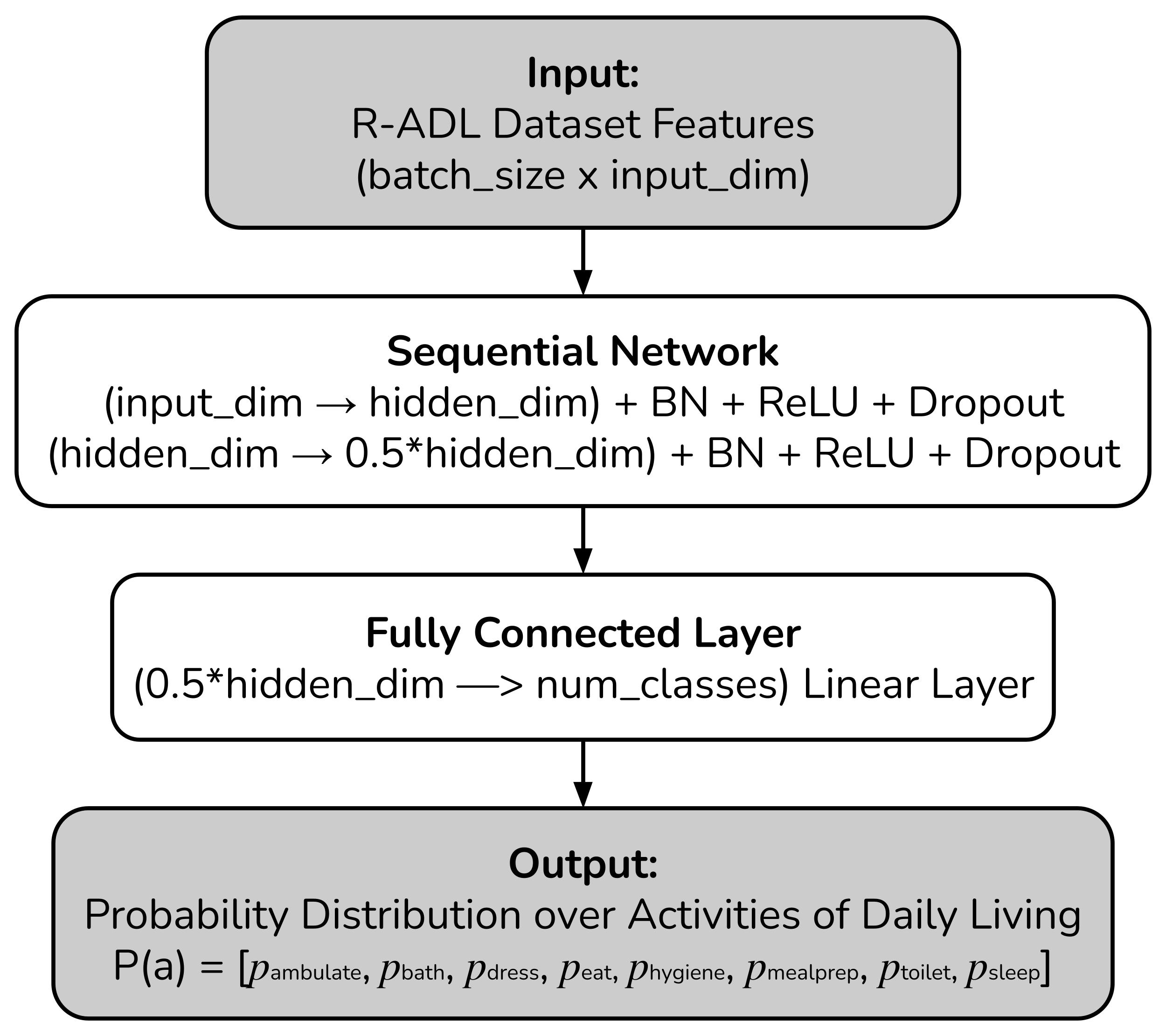}
    \caption{The neural network architecture for ADL classification. White boxes show layers of the network and gray boxes show data inputs and outputs.}
    \label{ADLNeuralNet}
\end{figure}

We use the \texttt{RandomForestClassifier} provided by \texttt{scikit-learn} to implement the random forest model, and the neural network model is implemented in \texttt{PyTorch}. The corresponding hyperparameters are summarized in Tables \ref{tab:RFParameters} and \ref{tab:TrainingADLNN}. While the random forest architecture is determined by the \texttt{RandomForestClassifier} implementation and its hyperparameters, the neural network architecture must be explicitly designed by specifying the arrangement of hidden layers and other network components. The resulting architecture is illustrated in Figure \ref{ADLNeuralNet}, which shows how Rectified Linear Unit (ReLU), Batch Normalization (BN), and Dropout operations are used to support prediction.

\begin{table}
\centering
\caption{Hyperparameters selected for the random forest model.}
\label{tab:RFParameters}

\begin{tabular}{|c|c|}
\hline
\multicolumn{1}{|c|}{\textbf{Parameter}} &
\multicolumn{1}{c|}{\textbf{Value}} \\ \hline

n\_estimators & 500 \\ \hline
class\_weight & balanced \\ \hline
max\_features & log \\ \hline
min\_samples\_split & 20 \\ \hline
min\_samples\_leaf & 10 \\ \hline

\end{tabular}
\end{table}

\begin{table}
\centering
\caption{Hyperparameters selected for the neural network model.}
\label{tab:TrainingADLNN}

\begin{tabular}{|c|c|}
\hline
\multicolumn{1}{|c|}{\textbf{Hyperparameter}} &
\multicolumn{1}{c|}{\textbf{Value}} \\ \hline

Hidden Dimension & 128 \\ \hline
Batch Size & 512 \\ \hline
Epochs & 40 \\ \hline
Patience & 5 \\ \hline
Learning Rate & 1e-3 \\ \hline
Optimizer & \texttt{AdamW} \\ \hline
Evaluation Criterion & \texttt{CrossEntropyLoss} \\ \hline
Model Selection Metric & \texttt{CrossEntropyLoss} \\ \hline
\end{tabular}
\end{table}

\subsection{Test Protocol}

In order to produce baseline results, we first split the R-ADL dataset into subsets for training and testing. We consider each time an activity was performed independently, splitting it to be included in both training and testing. This prevents class imbalance in training and testing, and ensures that both datasets exhibit similar data distributions.

\begin{figure}
    \centering
    \includegraphics[width=\columnwidth]{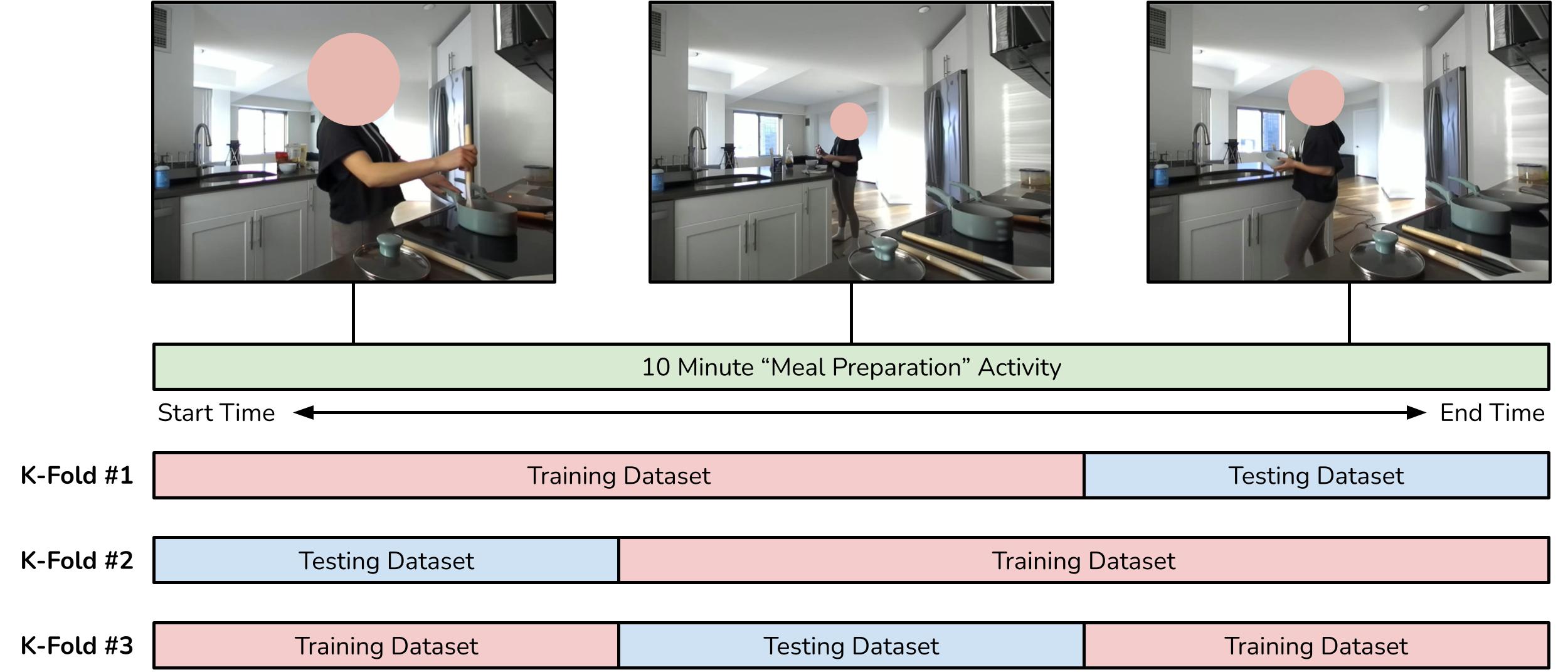}
    \caption{A visualization of the k-fold cross validation for baseline evaluation.}
    \label{KFoldScalability}
\end{figure}

For each activity, we use k-fold cross validation to split our data into three separate sections. This allows us to produce three different results for each model by assigning one of the sections to the testing split, while the other two are combined in the training split. This process is shown in Figure \ref{KFoldScalability}. This analysis technique reduces variance in the results and improves confidence in the conclusions made from performance outputs. We report averaged k-fold accuracy, F1 score, precision, and recall to summarize model performance in the ADL classification task. These performance metrics are collectively well-suited towards evaluation of multi-class classification problems. We provide precision, recall, and F1 score as macro-averaged across activity classes.

\subsection{Results}

\begin{figure}
    \centering
    \includegraphics[width=0.7\columnwidth]{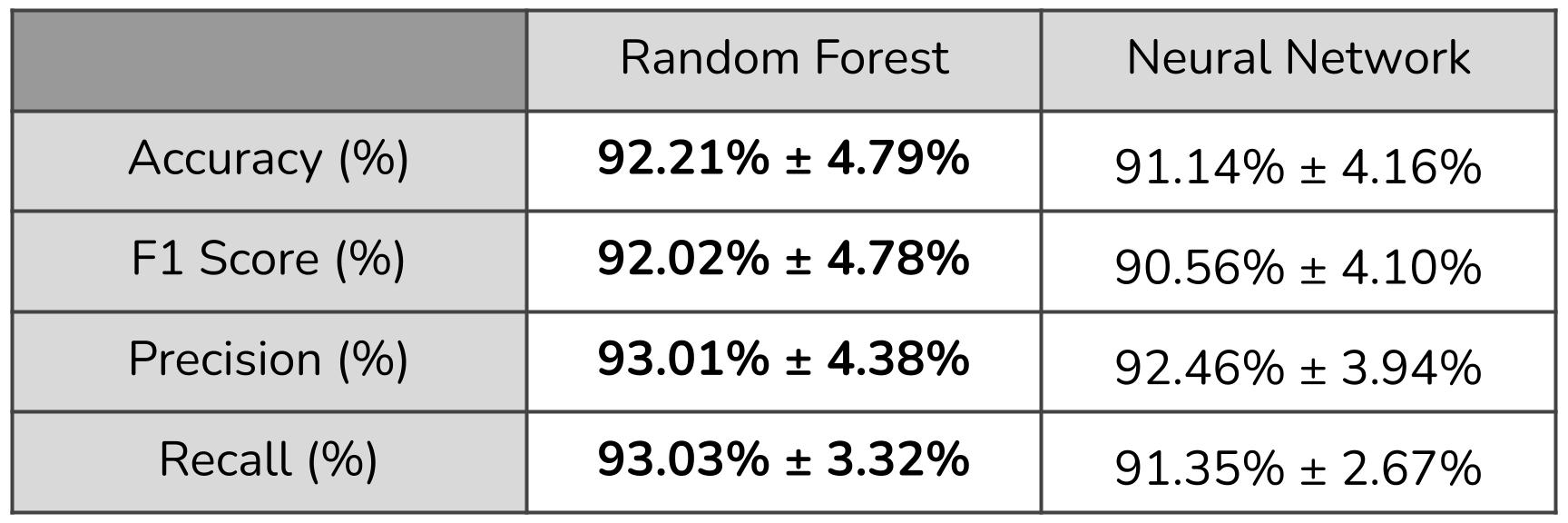}
    \caption{The baseline results for the random forest and neural network models when performing ADL classification on the R-ADL dataset. Standard deviation is reported across the three cross-validation folds.}
    \label{BaselinePerformance}
\end{figure}

Figure \ref{BaselinePerformance} shows the baseline results for the random forest and neural network models when performing ADL classification on the R-ADL dataset. Standard deviation is reported across the three cross-validation folds. This baseline demonstrates that both methods produce reliable ADL predictions, achieving over 90\% performance across all evaluation metrics. This establishes a foundation for future work to extend and enhance predictive capabilities using the R-ADL dataset.

\section{Contributions and Future Work}

This paper makes the following contributions to the field of activity recognition research:

\begin{itemize}
    \item We introduce the Radar-based Activities of Daily Living (R-ADL) dataset, a novel contribution for 1) its use of privacy-preserving sensors and 2) its intentional design to overcome the limitations of prior work in scale, environmental diversity, and support for high-level activities.
    \item We present a comprehensive analysis of how ADLs are prioritized across academic research, health survey tools, and clinical trials, providing evidence-based guidance for picking activities relevant to healthcare and senior living.
    \item We document the methodology underlying the development of the R-ADL dataset, including the experimental design, data collection protocol, and annotation strategy, ensuring reproducible data collection for future research.
    \item We establish baseline performance on representative machine learning models to provide a standardized point of comparison for new methods evaluated with R-ADL.
\end{itemize}

Future work should continue to expand the size and diversity of the R-ADL dataset. In particular, greater emphasis should be placed on recruiting a larger and more diverse participant population, especially older adults, who represent the primary target population for healthcare monitoring applications. Data collection should also be extended to additional homes and multi-person scenarios to improve model generalization across diverse living environments. Ultimately, the greatest value would come from collecting data in participants' own homes during their daily routines rather than through scripted activity sessions. Such data would enable the study of temporal patterns in daily living, including the timing, frequency, and sequencing of activities, while providing a more realistic representation of everyday behavior. 

The R-ADL dataset establishes a foundation for radar-based activity recognition research by providing both a large-scale, clinically motivated dataset and a reproducible methodology for dataset development. It serves as a valuable resource for advancing robust, privacy-preserving ADL recognition systems and we hope it will inspire future efforts toward larger, more diverse, and increasingly representative radar datasets.

\bibliographystyle{IEEEtran}
\bibliography{output}

\end{document}